\documentclass[conference]{IEEEtran}

\usepackage[utf8]{inputenc}
\usepackage{amsmath,amssymb,amsfonts}
\usepackage{cite}
\usepackage{graphicx}
\usepackage{textcomp}
\usepackage{microtype}
\usepackage[hidelinks]{hyperref}
\usepackage{booktabs}
\usepackage{csvsimple}
\usepackage[dvipsnames]{xcolor}
\usepackage{multirow}

\def\BibTeX{{\rm B\kern-.05em{\sc i\kern-.025em b}\kern-.08em
    T\kern-.1667em\lower.7ex\hbox{E}\kern-.125emX}}

\begin{document}

\title{Modeling Non-deterministic Human Behaviors in Discrete Food Choices}

\author{
    \IEEEauthorblockN{Andrew Starnes\IEEEauthorrefmark{1},
                     Anton Dereventsov\IEEEauthorrefmark{1},
                     E. Susanne Blazek\IEEEauthorrefmark{2},
                     Folasade Phillips\IEEEauthorrefmark{2}}
    \IEEEauthorblockA{\IEEEauthorrefmark{1}\textit{Lirio AI Research, Lirio LLC}, Knoxville, TN, USA}
    \IEEEauthorblockA{\IEEEauthorrefmark{2}\textit{Behavioral Reinforcement Learning Lab, Lirio LLC}, Nashville, TN, USA}
    \{astarnes, adereventsov, sblazek, sphillips\}@lirio.com
}

\maketitle

%%%%%%%%%%%%%%%%%%%%%%%%%%%%%%%%%%%%%%%%%%%%%%%%%%%%%%%%%%%%%%%%%%%%%%%%%%%%%%%%%%%%%%%%%%%%%%%%%%%
\begin{abstract}
We establish a non-deterministic model that predicts a user's food preferences from their demographic information.
Our simulator is based on NHANES dataset and domain expert knowledge in the form of established behavioral studies.
Our model can be used to generate an arbitrary amount of synthetic datapoints that are similar in distribution to the original dataset and align with behavioral science expectations.
Such a simulator can be used in a variety of machine learning tasks and especially in applications requiring human behavior prediction.
\end{abstract}

%%%%%%%%%%%%%%%%%%%%%%%%%%%%%%%%%%%%%%%%%%%%%%%%%%%%%%%%%%%%%%%%%%%%%%%%%%%%%%%%%%%%%%%%%%%%%%%%%%%

%%%%%%%%%%%%%%%%%%%%%%%%%%%%%%%%%%%%%%%%%%%%%%%%%%%%%%%%%%%%%%%%%%%%%%%%%%%%%%%%%%%%%%%%%%%%%%%%%%%
\section{Introduction}
Hyperpersonalization and nudging applications are frequently laced with problems caused by the non-deterministic nature of human behavior.
Self-report data in particular often runs the risk of producing unclear or inaccurate results, as an individual's perception of their own behaviors may vary drastically from one day to the next due to an infinite number of unknown external factors.
As a result, it becomes difficult to decipher whether stochasticity in self-report results is due to the input features themselves, or some sort of unknown, outside influences that caused a participant to recall a distorted view of their behaviors.
Being able to account for and adapt to this uncertainty is crucial for the development of resilient personalization approaches in a number of real-life applications, see e.g.~\cite{mills2022personalized}.

This paper addresses the issue of modeling human behavior based on observations collected from survey questionnaires, which is a particularly challenging task due to the fundamental stochasticity of the given data.
In addition to the non-deterministic nature of human behavior, this data collection via surveys often introduces additional noise to the recorded responses.

Learning stochastic processes with conventional data approximation methods~--- e.g. statistical inference, dictionary decomposition, machine learning, etc.~--- is known to be a highly challenging task, see e.g.~\cite{angluin1988learning, natarajan2013learning, raychev2016learning, aswani2018inverse}.
Typically in cases where the stochasticity is sufficiently low, the given data is treated as a perturbation of the underlying ``ground truth.''
While such an approach is widely utilized in some practical applications (such as image segmentation, face recognition, asset pricing, etc.) it is not a universal solution, as it fails to generalize to all real-world scenarios.

In particular, the issue of human feedback is not handled well by such methods since ``human-level stochasticity'' generally cannot be viewed as a slight perturbation of the underlying manifold.
For instance, consider a user leaving feedback (like/dislike) on a piece of media content.
The outcome of this interaction depends on a number of external factors that affect the mood of the user, which in turn affect the user's rating.
Thus, fully predicting the feedback of a user is generally an unfeasible task since the same user might provide different ratings to the same piece of content under the influence of different external factors.

This issue particularly plagues machine learning algorithms, as they are often designed to treat the training data as ground truth.
Indeed, a typical machine learning algorithm assumes that human behavior, or at least the feedback received from human behavior, is deterministic.
Even though there has been a lot of development in robust machine learning (see e.g.~\cite{song2022learning} and the references therein), when faced with fundamentally stochastic data, most conventional approaches perform unsatisfactorily~\cite{erev2017anomalies}.

In this paper we propose a method of modeling human behavior data by using statistical modeling and further actualize the predictions by implementing a domain expert knowledge in the form of behavioral science research that is relevant to the setting at hand.

%%%%%%%%%%%%%%%%%%%%%%%%%%%%%%%%%%%%%%%%%%%%%%%%%%%%%%%%%%%%%%%%%%%%%%%%%%%%%%%%%%%%%%%%%%%%%%%%%%%
\subsection{Behavioral Science and Behavior Modeling}
Behavioral science is a multidisciplinary field that leverages psychology, sociology, economics, and other various disciplines to understand human behavior and how individuals make decisions in the real world.
It functions as a diverse field of study with numerous applications in areas such as public health, finance, and marketing. 
Behavioral research serves as a method to examine and understand a multitude of human behaviors and the different factors that influence our decisions, attitudes, and beliefs.
Behavioral scientists conduct research and run experiments to better understand human actions and why people do the things they do.
In the social sciences, meta-analyses are becoming increasingly popular as methods for examining data from a number of independent studies of the same subject in order to determine overall trends.
Meta-analyses involve statistical analysis that combines the results of multiple scientific studies.
The accumulation of results across studies is the only known solution to the problem of sampling error inherent in small sample studies~\cite{hunter1991meta}.
Advantages of meta-analysis include, among others, illuminating trends across independent studies, maintaining statistical significance, minimizing wasted data, and finding moderator variables~\cite{rosenthal2001meta}.
Further improving meta-analytic methods by improving the transparency and reproducibility of meta-analyses is becoming a priority~\cite{lakens2017examining}.
Since 2018, more than 32,500 meta-analyses in behavioral science have been made available (Google Scholar search of “meta-analysis” + “behavior” + “choice”). 

Our main goal in building a stochastic model of human feedback is to evaluate the extent to which identical demographic inputs can result in different outcomes in food preferences, due to the stochasticity and lack of predictability inherent in each of the demographics (as identified by the behavioral science literature).
Variations in human preferences are due to individual factors (e.g., age, self-efficacy), external factors (e.g., context, day), engagement factors (e.g., exposures over time), and more.
Every single factor plays a role in the stochasticity of human preferences, but we cannot know what all the factors are nor the size of the role(s) they play.
In this work we model human feedback as food preferences, with demographic variables serving as the sources of stochasticity.

%%%%%%%%%%%%%%%%%%%%%%%%%%%%%%%%%%%%%%%%%%%%%%%%%%%%%%%%%%%%%%%%%%%%%%%%%%%%%%%%%%%%%%%%%%%%%%%%%%%
\subsection{Related Work}
The focus of this paper is to enhance the conventional statistical learning approach by employing domain expert knowledge in the form of behavioral science research.
Such an approach is regularly considered in relevant literature, see e.g.~\cite{peysakhovich2017using, turgeon2020tutorial, hagen2020can, mac2020artificial}.

We are addressing the task of modeling human behavior.
Such a task is considered in Choice Prediction Competition\footnote{\url{https://cpc-18.com/}} and its follow-up literature, e.g.~\cite{erev2010choice, plonsky2018and, bourgin2019cognitive}.

We are using NHANES dataset to learn food preferences from the given demographic data.
This dataset is regularly used in behavioral literature, see e.g.~\cite{zadshir2005prevalence, block2004foods, mozumdar2011persistent, storz2021diet, storz2022shiftwork}.

We leverage behavioral science studies to construct a non-deterministic model of human behavior. 
A similar approach is taken in~\cite{capote2012stochastic}, where the authors propose a non-deterministic model of human behavior during train evacuation processes; however, they do not take into account any of the individuals' information, and thus their model predicts an average behavior rather than that of a given person.

In particular, we use the Theory of Planned Behavior (TPB) to model non-determinicity of human decisions.
Similar approach was utilized in~\cite{dunn2011determinants, ajzen2015consumer}, where the authors employed the TPB in the context of food preferences.

%%%%%%%%%%%%%%%%%%%%%%%%%%%%%%%%%%%%%%%%%%%%%%%%%%%%%%%%%%%%%%%%%%%%%%%%%%%%%%%%%%%%%%%%%%%%%%%%%%%
\section{Problem Formulation}
We construct a non-deterministic model to approximate human decision-making on an example of food choice selection.
Specifically, we utilize NHANES dataset, examined in Section~\ref{sec:nhanes}, to model the participants' food choices based on their demographic information.
We use the dataset to train a statistical model as stated in Section~\ref{sec:data_modeling} and employ the behavioral insights outlined in Section~\ref{sec:besci_insights} and summarized in Table~\ref{tab:besci_metaanalyses}.

%%%%%%%%%%%%%%%%%%%%%%%%%%%%%%%%%%%%%%%%%%%%%%%%%%%%%%%%%%%%%%%%%%%%%%%%%%%%%%%%%%%%%%%%%%%%%%%%%%%
\subsection{NHANES Dataset}\label{sec:nhanes}
In this paper we utilize data from the Centers for Disease Control and Prevention’s National Health and Nutrition Examination Survey (NHANES) repository~\cite{nhanes}.
We extract participants' demographic features and food preferences from NHANES 2017--2018 survey\footnote{\url{https://wwwn.cdc.gov/nchs/nhanes/continuousnhanes/default.aspx?BeginYear=2017}} to construct a non-deterministic model that predicts the participants' food choice selections based on their demographic features.
NHANES is a large, biennial, stratified, multistage survey conducted by the Centers for Disease Control and Prevention and is one of the largest and most important cross-sectional studies conducted in the United States in terms of participant size, scope, ethical diversity, and free data accessibility~\cite{leroux2019organizing}.

For this study, we use data from various modules within the NHANES, including demographic and questionnaire data. 
Specifically, we use the Diet Behavior and Nutrition questionnaire (DBQ) that asks participants about their eating styles over the past week, available for download at~\url{https://wwwn.cdc.gov/Nchs/Nhanes/2017-2018/P_DBQ.XPT} with a detailed description of the data collection at~\url{https://wwwn.cdc.gov/nchs/data/nhanes/2017-2018/questionnaires/DBQ_J.pdf}.
The data contains 29 demographic and 46 response features for the 15,560 participants.
Demographic data includes gender, age, marital status, race/ethnicity, education level, and household income.
Each participant is asked how many times they ate out (fast-food, restaurants, vending machines, etc.) vs how many times they ate at home, out of total $21$ meals (assuming breakfast, lunch, and dinner every day).

%The first questions inquires about the number of meals not prepared at home during the past 7 days (e.g., meals purchased in restaurants or fast-food places or obtained from vending machines).
%Two additional questions ask for the number of ready-to-eat foods (e.g., store prepared soups, chicken, and sandwiches) and the number of frozen meals/pizzas eaten in the past 30 days.
%
%From the question descriptions alone it is unclear whether the three food choices are entirely independent, i.e., whether the number of ready-to-eat foods and the number of frozen meals/pizzas eaten in the past 30 days are a subset of the number of meals not prepared at home during the past 7 days.
%Recent publications using these three NHANES questions assume the three food choices were independent, or at least do not mention otherwise~\cite{storz2021diet, storz2022shiftwork}.

%%%%%%%%%%%%%%%%%%%%%%%%%%%%%%%%%%%%%%%%%%%%%%%%%%%%%%%%%%%%%%%%%%%%%%%%%%%%%%%%%%%%%%%%%%%%%%%%%%%
\subsection{Behavioral Expert Knowledge}\label{sec:besci_insights}
The Theory of Planned Behavior (TPB) has been the behavioral science theory of study in at least 56 meta-analyses (Google Scholar search of published article titles including all the words “meta-analysis” + “Theory of Planned Behavior”). TPB, an extension of the Theory of Reasoned Action~\cite{ajzen1985intentions}, is a psychological theory developed by Icek Ajzen~\cite{fishbein1977belief} as a model to predict and understand an individual's behavior.
The TPB argues that an individual's behavioral intention is collectively shaped and influenced by three main components: an individual’s attitude, societies’ subjective norms, and their perceived control of the behavior~\cite{ajzen1985intentions}.
TPB has been leveraged to successfully predict and explain a variety of health-related behaviors from physical activity to smoking~\cite{wing2009predicting, norman1999theory}.
Food choice is a complex human behavior that is influenced by a wide range of individual and environmental factors.

For the purposes of this paper, we choose to leverage the TPB as this theory offers a strong theoretical basis for trying to understand the role and influence of certain factors on health-related behaviors like food choice.
Namely, we utilize the available behavioral research (listed below) for the demographic data used in our study (gender, age, marital status, race/ethnicity, education level, and household income).

\subsubsection{Gender}
Research has highlighted the relationship between gender and dietary choices.
For example, in a study that examined gender differences in fruit and vegetable consumption through the lens of the TPB, it was found that women were more likely to consume fruits and vegetables when compared to men~\cite{emanuel2012theory}.

\subsubsection{Age}
Age predicts dietary preferences and choices.
For example, teenagers and younger adults are less likely to consume health promoting foods~\cite{mcdermott2015theory}.
A systematic review and meta-analysis on the association between variables specified by the TPB (i.e., Attitude, Social Norms, Perceived Behavioral Control) and discrete food choice behaviors showed stronger pooled mean effect sizes of the association between intentions and behaviors as age went up.
Being in the age group 18-29 was associated with 1) less frequent deviation from a choice (i.e., stronger intention-behavior association) and 2) when there was deviation, less of it (i.e., lower heterogeneity).

\subsubsection{Marital Status}
Some research has explored the role of marital status in dietary behaviors and food choices~\cite{konttinen2021sociodemographic, lee2013understanding}.
In a study that was conducted to understand the influence of sociodemographic factors on an individual’s healthy eating behaviors in restaurants using both the TPB and the Health Belief Model, it was shown that singles were more willing than married couples to patronize a restaurant that offered healthy food choices~\cite{lee2013understanding}.
Marital status comprised the following categories: married or living with a partner, widowed/divorced/separated, and never married.

\subsubsection{Race/Ethnicity}
Increasing evidence indicates that race plays a critical role in diet-related disparities, with racial and ethnic minority groups (i.e. African American, Hispanic, American Indian/Alaska natives) typically having poorer nutrient profiles and dietary behaviors/patterns compared to the majority population groups (e.g., white/European Americans~\cite{satia2009diet}).
Explanations for this phenomenon can be attributed to social disadvantage, and it is well documented that minority racial groups tend to have higher levels of unemployment and lower income~\cite{barr2014health}. 

\subsubsection{Education}
Prior research has explored the relationship between educational attainment and diet quality, showing individuals with higher educational attainment adhering to healthier diets~\cite{backholer2016association, darmon2015contribution}. 

\subsubsection{Household Income}
Nutrient-rich food tends to cost more per calorie compared to food with lower nutritional values~\cite{darmon2015contribution}.
Individuals from high socioeconomic status (SES) groups tend to consume healthier and more expensive diets and individuals from low SES groups tend to select cheaper, more energy dense diets lacking in fruits and vegetables.
Lower SES is associated with less disposable income, thus creating a barrier to adhering to a healthy diet~\cite{darmon2015contribution, bertoni2011multilevel}.

\begin{table*}
    \centering
    \caption{Demographic Determinants of Choice Uncertainty}
    \begin{tabular}{|c|c|c|c|}
        \hline
        \textbf{Demographic} & \multicolumn{3}{|c|}{\textbf{Dichotomized Choice Uncertainty}}
        \\\cline{2-4}
        \textbf{Determinants} & \textbf{\textit{Direction}} & \textbf{\textit{Strength}} & \textbf{\textit{Meta-analyses}}
        \\\hline
        Gender & \multirow{2}{*}{Positive} & \multirow{2}{*}{Very small} & \multirow{2}{*}{\cite{mcdermott2015theory, mcdermott2015theory2, riebl2015systematic, steinmetz2016effective}}
        \\(male vs. female) &&&
        \\\hline
        Age & \multirow{2}{*}{Negative} & \multirow{2}{*}{Small} & \multirow{2}{*}{\cite{mcdermott2015theory, mcdermott2015theory2, mceachan2011prospective}}
		\\(high vs. low) &&&
        \\\hline
        Marital status & \multirow{2}{*}{Positive} & \multirow{2}{*}{Small} & \multirow{2}{*}{\cite{lee2013understanding}$^{\mathrm{a}}$}
        \\(married vs. single) &&&
        \\\hline
        Race/ethnicity & \multirow{2}{*}{None} & \multirow{2}{*}{None} & \multirow{2}{*}{\cite{li2019socioeconomic}}
        \\(majority vs. minority) &&&
        \\\hline
        Education & \multirow{2}{*}{Positive} & \multirow{2}{*}{Very small} & \multirow{2}{*}{\cite{steinmetz2016effective, li2019socioeconomic}}
        \\(high vs. low) &&&
        \\\hline
        Household income & \multirow{2}{*}{None} & \multirow{2}{*}{None} & \multirow{2}{*}{\cite{li2019socioeconomic}}
        \\(high vs. low) &&&
        \\\hline
        \multicolumn{4}{l}{$^{\mathrm{a}}$Not a meta-analysis. None were found that explored the effect of marital status.}
    \end{tabular}
    \label{tab:besci_metaanalyses}
\end{table*}

We are interested in modeling choice uncertainty as it relates to food choices, where uncertainty is how often does someone deviates from a choice. See Table~\ref{tab:besci_metaanalyses} for a summary of demographic determinants of choice uncertainty.
In the provided table, the ``Direction'' refers to whether a positive or negative association was found between demographic variables and uncertainty in food choices.
``Strength'' refers to the magnitude of the positive or negative association found between demographic variables and uncertainty in food choices.
And ``None'' indicates no moderating effects of demographics on healthy food choices were found, or that the literature was sufficiently mixed to warrant an absence of definitive direction and strength.

%%%%%%%%%%%%%%%%%%%%%%%%%%%%%%%%%%%%%%%%%%%%%%%%%%%%%%%%%%%%%%%%%%%%%%%%%%%%%%%%%%%%%%%%%%%%%%%%%%%
\section{Modeling Food Choices}\label{sec:data_modeling}
We begin this section by motivating the need for a behavioral data simulator.
If someone is deciding whether or not to eat at home, then there is some probability that at that particular instant in time they might eat at home.
Denote this probability as $p\in[0,1]$.
Let $X\sim Bern(p)$ (i.e., $\mathbb{P}(X=1)=p$ and $\mathbb{P}(X=0)=1-p$) which represents the person's decision of where to eat; that is, if $X=0$ then the person will eat at home and if $X=1$ then the person eats elsewhere.

In our dataset, we have the number of times that each person ate somewhere other than home over 7 days (the corresponding probability mass function will be referred to as the ``population PMF'').
We will assume that everyone made 3 decisions on where to eat each day, for a total of 21 decisions.
Let $x_{n,i}$ represent person $n$'s decision for the $i$-th meal.
So, $x_{n,i}=1$ means the person did not eat at home and $x_{n,i}=0$ means the person are at home or did not eat meal $i$.
These are samples from $X_{n,i}\sim Bern(p_{n,i})$, where $p_{n,i}$ is the likelihood of person $n$ eating out for meal $i$.
Let $Y_n$ represent the number of times person $n$ ate out over these 21 meals.
This means $Y_n=\sum_{i=1}^{21}X_{n,i}$.

Modeling the number of times each person will eat out boils down to modeling each of their meal decisions.
In the dataset, we only have one observation for each person, which makes it impossible to accurately model the probabilities $p_{n,i}$ for each person.
Instead we will model the ``average'' person and then we will modify the probabilities using either the data or expert knowledge.
Modifications can be done as follows: if we estimate the $X_i\sim Bern(p_i)$ for meal $i$, we can regress $p_i$ to or away from 0.5 if there is more or less uncertainty in this person's choices based on their demographics, respectively.

In order to simulate the data, we begin by identifying $M$ peaks and plateau-like events in the population PMF (denoted by $h$), call them $\{k_1,...,k_M\}$.
In particular, $M$ is chosen to allow the simulated distribution to approximate the population distribution.
We will create $M$ Binomial random variables, $X_1,...,X_M$, where $X_i\sim Binom(n_i,p_i)$ which is the sum of $n_i$ i.i.d. $Bern(p_i)$ random variables.
Next, we specify the variance of $X_i$ for each $i$, denoted $\sigma_i^2$.
Recall that if $X\sim Binom(n,p)$, then $E(X)=np$, $Var(X)=np(1-p)$, and the maximum of the PMF is at $np$.
We use these facts to specify that $n_ip_i=k_i$ and $n_ip_i(1-p_i)=\sigma_i^2$.
This specifies
\[
    p_i=1-\frac{\sigma_i^2}{k_i}
    \text{ and }
    n_i=\frac{k_i}{p_i}.
\]
Finally, we want to add together the PMFs of $X_1,...,X_M$ in a way so that the sum is also a PMF.
We can use linear regression to solve the following system, which is almost certainly over-specified:
\begin{align*}
    \sum_{j}w_j\binom{n_j}{k_i}p_j^{k_i}(1-p_j)^{n_j-k_i}&=h(k_i)
    \quad\forall i
    \\
    \sum_{i}w_i&=1
\end{align*}
Let $Y=\sum_{i}w_iX_i$, then $Y$ is a random variable that approximates the population's distribution of counts.

We can modify the probabilities of $X_i$ based on the uncertainty of someone's decision using expert knowledge and, in turn, this modifies the distribution of $Y$.
Let $\alpha\in (0,1)$ represent the uncertainty change.
For each $i$, $\hat{p}_i = (1\pm\alpha)\cdot(p_i-0.5)+0.5$ where a ``$+$'' indicates less uncertainty and a ``$-$'' indicates more uncertainty (obviously constraining $\hat{p}_i$ to be in $[0,1]$).
For example, according to Table~\ref{tab:besci_metaanalyses}, males have more uncertainty in decisions than females.
Let $\hat{p}^{m}_i = (1-\alpha)\cdot(p_i-0.5)+0.5$, $\hat{p}^{f}_i = (1+\alpha)\cdot(p_i-0.5)+0.5$, $X^m_i\sim Binom(n_i,\hat{p}^{m}_i)$ and $X^f_i\sim Binom(n_i,\hat{p}^{f}_i)$, where ``$m$'' indicates male and ``$f$'' indicates female.
Then $Y^{m}=\sum_{i}w_iX^m_i$ and $Y^{f}=\sum_{i}w_iX^f_i$ can be used to represent the counts for males and females, respectively.
Furthermore, we can use $X^{\cdot}_i$ or the corresponding Bernoulli random variables to simulate actual decisions.

%%%%%%%%%%%%%%%%%%%%%%%%%%%%%%%%%%%%%%%%%%%%%%%%%%%%%%%%%%%%%%%%%%%%%%%%%%%%%%%%%%%%%%%%%%%%%%%%%%%
\section{Experiment Setup}
In this section we present our non-deterministic simulator that predicts a person's food preferences from their demographic features.
\footnote{All relevant code can be found at \url{https://github.com/acstarnes/wain2022_food_choice_modeling}.}
The demographic information is given as the participant's gender, age, marital status, race/ethnicity, education, and income.
Food preferences are represented by the number of meals the participant ate out versus the number of meals the participant ate at home.

We first extract the relevant data from NHANES dataset, as outlined in Section~\ref{sec:nhanes} and preprocess the obtained data in the following way:
\begin{itemize}
	\item Continuous demographic features (age, household income) are normalized to be in $[0,1]$;
	\item Categorical demographic features (gender, race/ethnicity, marital status) are parameterized via one-hot encoding;
	\item Ordinal demographic features (education) are assigned to equidistant points in $[0,1]$;
    \item The food preference vector is normalized to form a probability distribution over the number of times chosen to eat out.
\end{itemize}
We drop any missing values from the original dataset and obtain a set of $7,784$ datapoints.
Training and test sets are created with a random 67\%-33\% split, respectively.
We then train a statistical model via the procedure stated in Section~\ref{sec:data_modeling} and further enhance the prediction results with the behavioral science insights discussed in Section~\ref{sec:besci_insights}.

\begin{figure}
    \centering
    \includegraphics[width=0.7\linewidth]{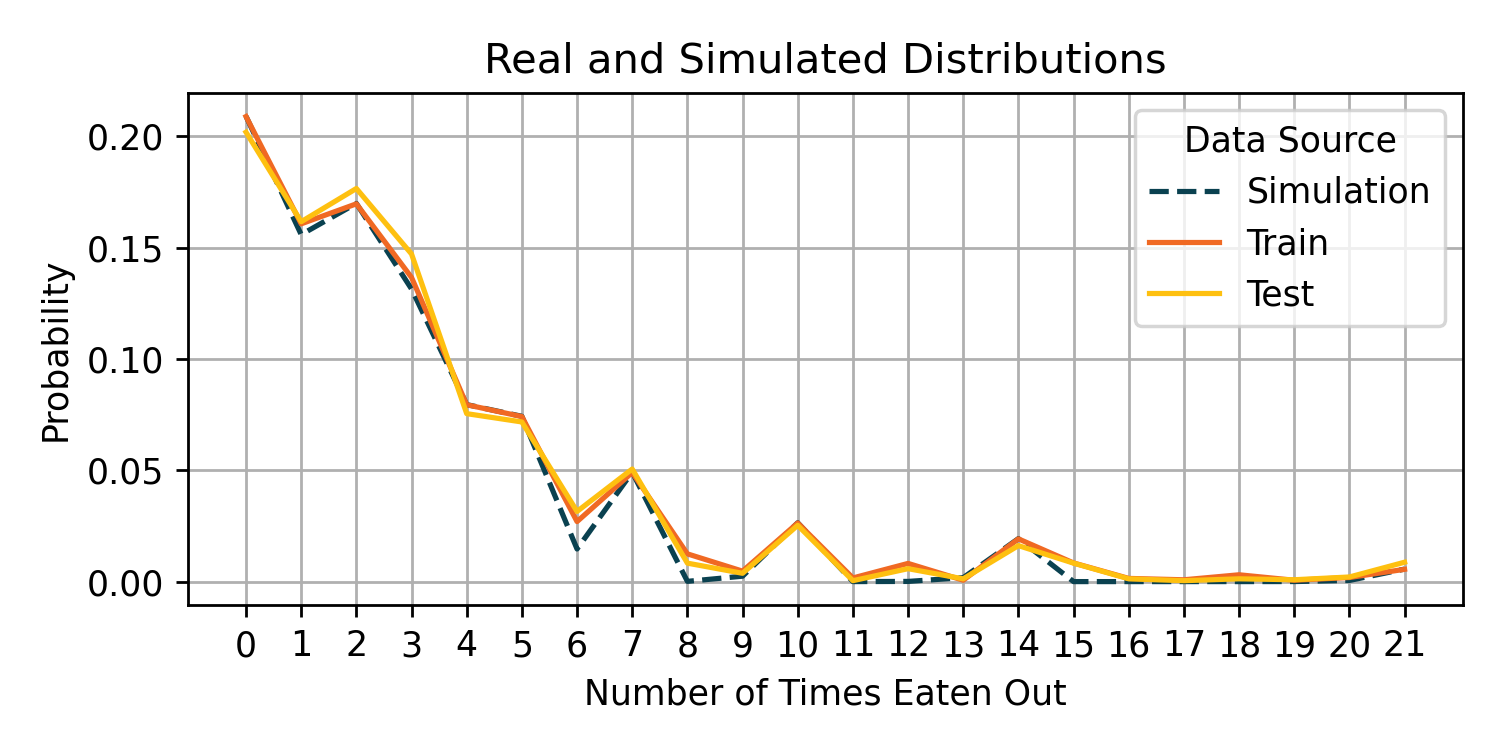}
    \caption{Probability mass distributions based on counts in train and test sets along with the simulated distribution}
    \label{fig:simulation}
\end{figure}

\begin{figure}
    \centering
    \includegraphics[width=0.8\linewidth]{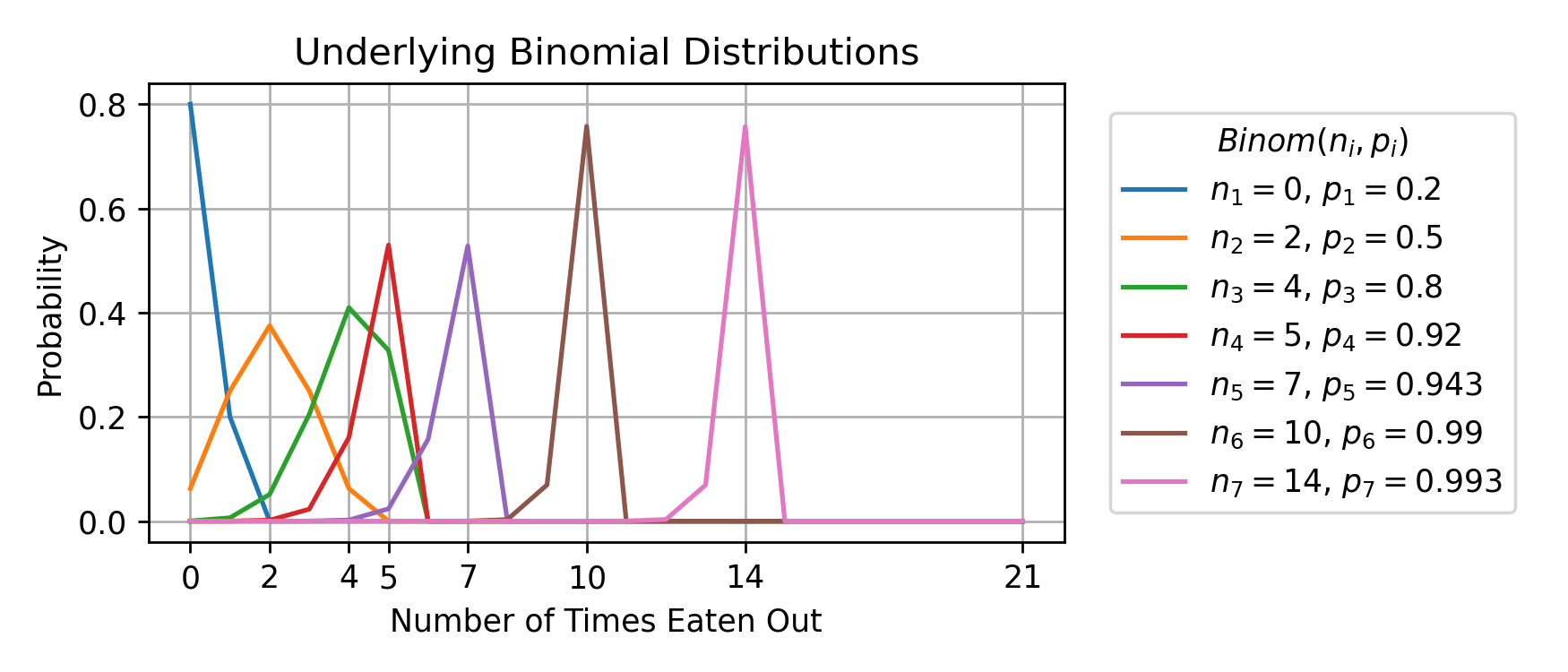}
    \caption{Binomial distributions centered around $k$ values with widths based on $\sigma^2$ values}
    \label{fig:binomial_distributions}
\end{figure}

We begin by plotting the probability distributions from the training and test sets in Figure~\ref{fig:simulation}, where we identify key points of support in the training distribution at $k_0=0$, $k_1=2$, $k_2=4$, $k_3=5$, $k_4=7$, $k_5=10$, $k_6=14$, and $k_7=21$.
We set the initial variances to 1 and modify them to more accurately match the training distribution.
Specifically, we use the following $\sigma_0^2=0.2$, $\sigma_1^2=1$, $\sigma_2^2=0.8$, $\sigma_3^2=0.1$, $\sigma_4^2=0.4$, $\sigma_5^2=0.1$, $\sigma_6^2=0.1$, $\sigma_7^2=0.1$.
Once we have the $k$'s and $\sigma^2$'s, we model the training distribution as in Section~\ref{sec:data_modeling}, which can be seen in Figure~\ref{fig:simulation}.
We use histogram intersection (HI) as our measure of accuracy and find that the training HI is $93.8\%$ and the test HI is $92.1\%$ (recall that HI values are between 0 and 1, with 1 indicating the distributions are the same).
The underlying binomial distributions, centered around the $k$ values, are shown in Figure~\ref{fig:binomial_distributions}.

\begin{figure}
    \centering
    \includegraphics[width=0.86\linewidth]{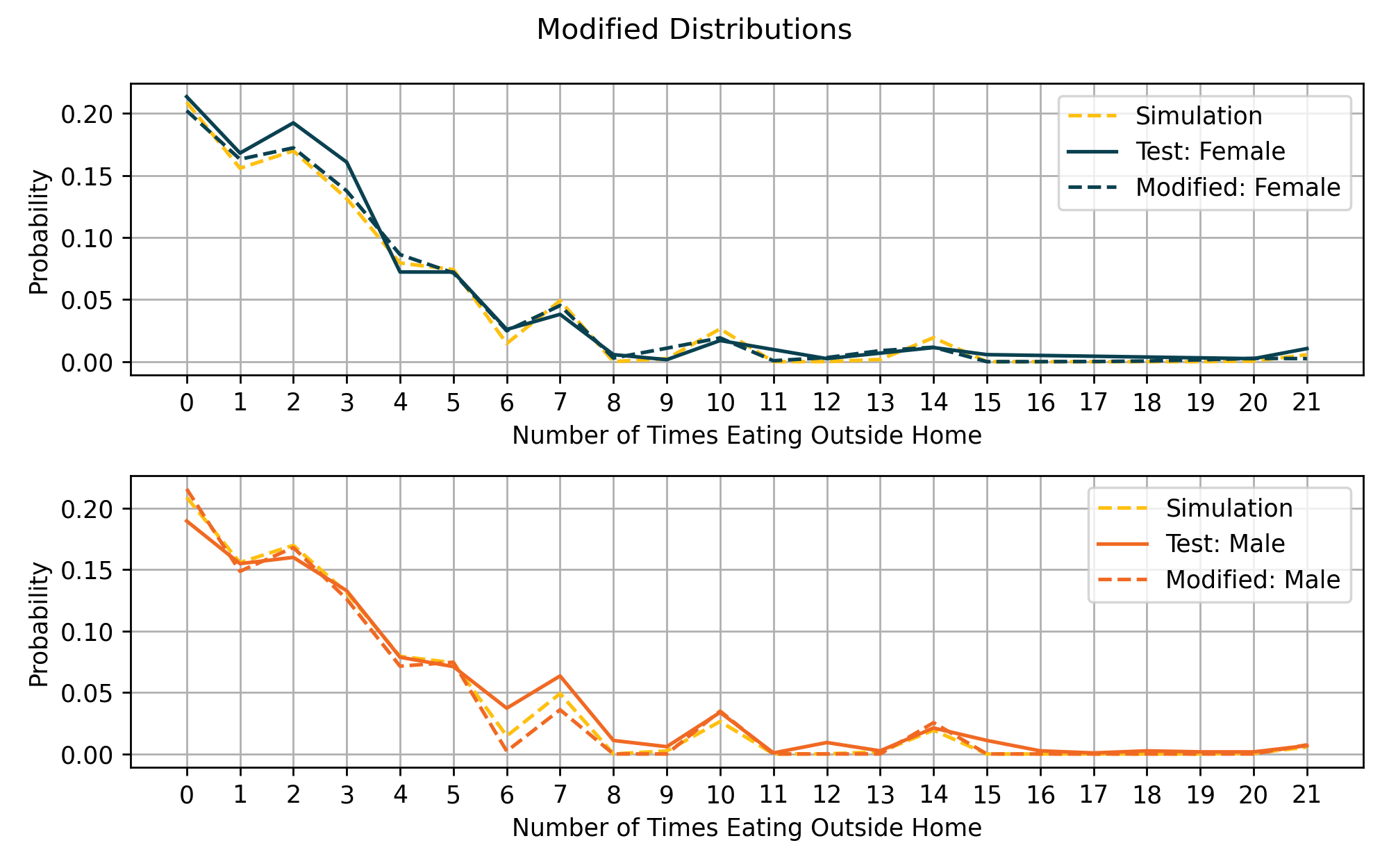}
    \caption{Modified distributions for males and females along with the corresponding distributions for the test sets}
    \label{fig:gender}
\end{figure}

\begin{figure}
    \centering
    \includegraphics[width=0.86\linewidth]{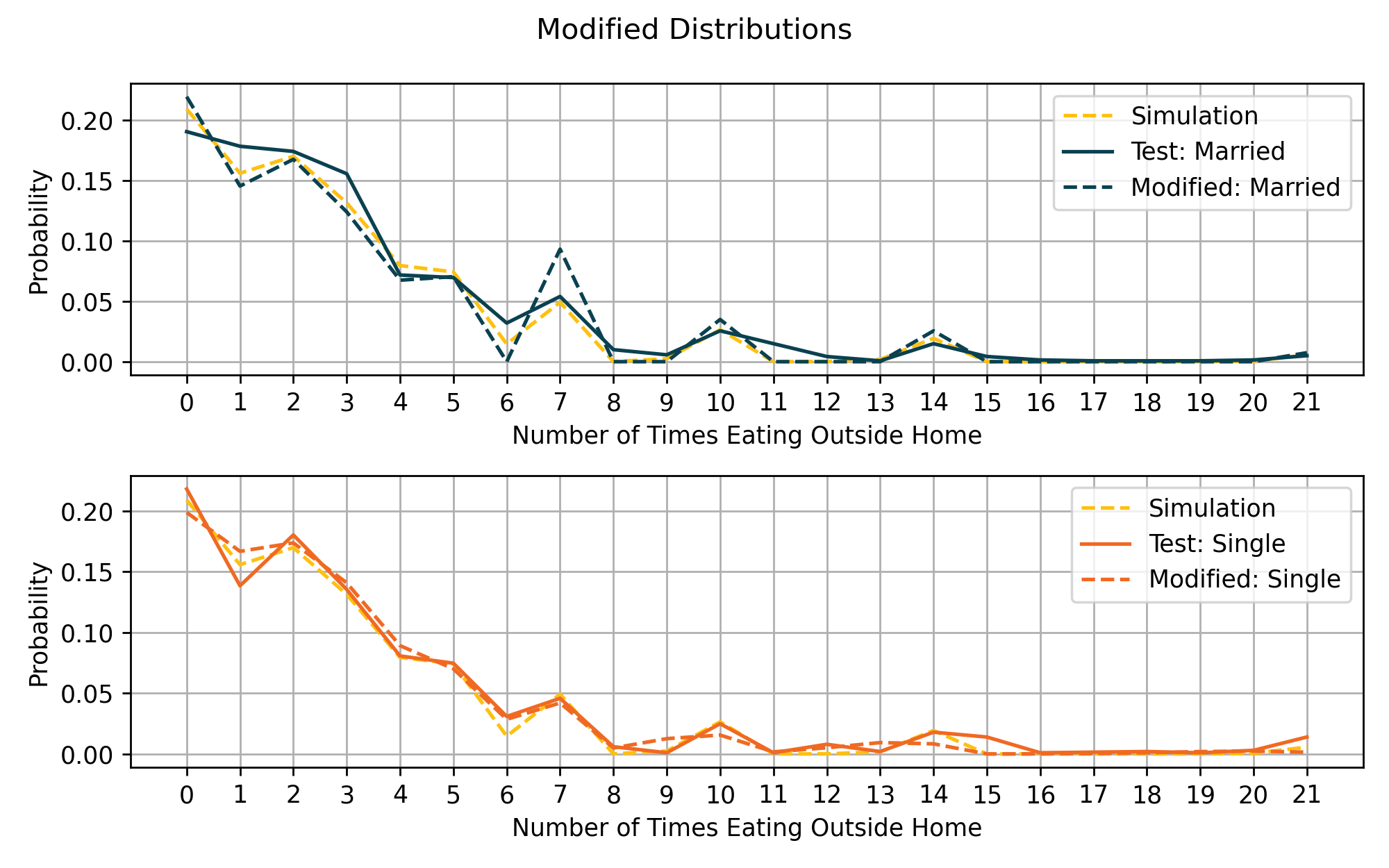}
    \caption{Modified distributions for marital status along with the corresponding distributions for the test sets}
    \label{fig:marital_status}
\end{figure}

We now will modify the binomial probabilities based on Table~\ref{tab:besci_metaanalyses}.
For gender, there is more uncertainty with males than with females. 
As such, we need to move all of the binomial probabilities closer to 0.5 for males and further from 0.5 for females.
Using a scaling factor of 0.1, the resulting probabilities for males are $p_0^m=0.23$, $p_1^m=0.5$, $p_2^m=0.77$, $p_3^m=0.878$, $p_4^m=0.899$, $p_5^m=0.941$, $p_6^m=0.944$, $p_7^m=0.946$ and for females are $p_0^f=0.17$, $p_1^f=0.5$, $p_2^f=0.83$, $p_3^f=0.962$, $p_4^f=0.987$, $p_5^f=1$, $p_6^f=1$, $p_7^f=1$.
Plotting the new distributions for male and female along with the distributions from the testing set can be seen in Figure~\ref{fig:gender}.
Additionally, the test accuracy for female is $87.4\%$ and for male is $91.0\%$.
We repeat this for marital status using a scaling factor of 0.15, the resulting distributions can be seen in Figure~\ref{fig:marital_status}.
Here, the test error when marital status is married is $86.3\%$ and when marital status is single is $91.2\%$.
The modifications made to the probabilities based on expert knowledge follow the true test distributions with reasonable accuracy.

Since the simulated distributions based on expert knowledge roughly follow the test distributions that the simulator was never trained on, data sampled from these distributions are reasonable approximations to the real data. We can use them to generate an unlimited number of new data points. One way to do this is simply by sampling from the distributions in order to obtain samples for gender and number of times eaten out. Alternatively, we can make 21 samples from the underlying binomial distributions in order to simulate individual meal choices and see the impact that gender has on them. Finally, we can simulate data based on features that are not available in our dataset by using expert knowledge and modifying the distributions as we did above with gender.

%%%%%%%%%%%%%%%%%%%%%%%%%%%%%%%%%%%%%%%%%%%%%%%%%%%%%%%%%%%%%%%%%%%%%%%%%%%%%%%%%%%%%%%%%%%%%%%%%%%
\section{Discussion}
In this paper we construct a non-deterministic model predicting human food preferences based on demographic information.
Our model is based on the combination of real data obtained from the NHANES dataset and the domain expert knowledge obtained from behavioral science studies.

Our model can be used as a behavioral data simulator to generate an arbitrary amount of food choice probabilities for given demographic features, either real or synthesized.
Such a simulator is of interest to machine learning practitioners who work with conventional approaches and thus require vast amounts of data unavailable in behavioral surveys.

While the current work is rather preliminary, it suggests a number of immediate promising directions by utilizing the data obtained from our simulator in machine learning applications dealing with behavioral data.
One area of particular interest in this direction is the training of reinforcement learning agents on feedback obtained from agent-human interactions, which, for instance, includes content recommendation algorithms.

\subsection*{Limitations}
Dichotomizing determinants of food uncertainty into low vs. high, positive vs. negative, etc. undermines the vast amount of behavioral science literature that repeatedly demonstrates the individual variation in human behavior.

The irony is not lost on these authors that our stochastic model, although a step up from previous deterministic models, is a general model that averages the wide variety of human behavior into a simplified reduction, all in the effort to train a reinforcement learning agent how to personalize.
Over time and with additional research, our goal is to increase the complexity of our stochastic model of human behavior in a way that better appreciates each individual’s variation in preferences.

It would not be feasible nor possible to hold all determinants constant, so even if we controlled all that we could humanly control, each individual will possibly make a different choice in the same context under the same circumstances every time.

%%%%%%%%%%%%%%%%%%%%%%%%%%%%%%%%%%%%%%%%%%%%%%%%%%%%%%%%%%%%%%%%%%%%%%%%%%%%%%%%%%%%%%%%%%%%%%%%%%%
\bibliographystyle{abbrv}
\bibliography{references}
%%%%%%%%%%%%%%%%%%%%%%%%%%%%%%%%%%%%%%%%%%%%%%%%%%%%%%%%%%%%%%%%%%%%%%%%%%%%%%%%%%%%%%%%%%%%%%%%%%%

\end{document}